\pdfoutput=1
\documentclass[11pt]{article}

\usepackage{authblk}

\usepackage{acl}

\usepackage{times}
\usepackage{latexsym}

\usepackage[T1]{fontenc}
\usepackage[utf8]{inputenc}

\usepackage{microtype}

\usepackage{graphicx}
\usepackage{color,soul}
\usepackage{makecell}
\usepackage{xcolor}
\usepackage{microtype}
\usepackage{enumitem}% http://ctan.org/pkg/enumitem
\usepackage{multirow}
\usepackage{booktabs}
\usepackage{etoolbox}
\usepackage{tabularx}
\usepackage[normalem]{ulem} % [normalem] prevents the package from changing the default behavior of `\emph` to underline.
\usepackage{relsize}

\AtBeginEnvironment{quote}{\par\singlespacing\small}
\newenvironment{myquote}%
{\list{}{\leftmargin=0.3in}\item[]}%
{\endlist}

\title{IDIAPers @ Causal News Corpus 2022: Extracting Cause-Effect-Signal Triplets via Pre-trained Autoregressive Language Model
}

\author[*, 1, 2]{\textbf{Martin Fajcik}}
\author[1]{\textbf{Muskaan Singh}}
\author[1,3]{\textbf{Juan Zuluaga-Gomez}}
\author[1,4]{\textbf{Esaú Villatoro-Tello}}

\author[1,5]{\authorcr\textbf{Sergio Burdisso}}
\author[1, 2]{\textbf{Petr Motlicek}}
\author[2]{\textbf{Pavel Smrz}}

\affil[1]{Idiap Research Institute, Martigny, Switzerland}
\affil[2]{Brno University of Technology, Brno, Czech Republic}
\affil[3]{Ecole Polytechnique Fédérale de Lausanne, Switzerland}
\affil[4]{Universidad Autónoma Metropolitana Unidad Cuajimalpa, Mexico City, Mexico}
\affil[5]{Universidad Nacional de San Luis (UNSL), San Luis, Argentina}
\affil[ ]{\emph{*corresponding author: martin.fajcik@vut.cz}}

\begin{document}
\maketitle
\begin{abstract}
In this paper, we describe our shared task submissions for Subtask 2 in CASE-2022, Event Causality Identification with Casual News Corpus. The challenge focused on the automatic detection of all cause-effect-signal spans present in the sentence from news-media. We detect cause-effect-signal spans in a sentence using T5 --- a pre-trained autoregressive language model. We iteratively identify all cause-effect-signal span triplets, always conditioning the prediction of the next triplet on the previously predicted ones. To predict the triplet itself, we consider different causal relationships such as \emph{cause$\rightarrow$effect$\rightarrow$signal}. Each triplet component is generated via a language model conditioned on the sentence, the previous parts of the current triplet, and previously predicted triplets.
Despite training on an extremely small dataset of 160 samples, our approach achieved competitive performance, being placed second in the competition. Furthermore, we show that assuming either \emph{cause$\rightarrow$effect} or \emph{effect$\rightarrow$cause} order achieves similar results.\footnote{Code at \url{https://github.com/idiap/cncsharedtask}.}
\end{abstract}

\section{Introduction}
Causality links the relationship between two arguments --- cause and effect \cite{barik2016event}. Figure \ref{Figure:example} shows  examples extracted from the Causal News Corpus (CNC) \cite{tan-EtAl:2022:LREC}. \textit{Cause} clauses appear in yellow, \textit{Effect} in green, and \textit{Signals} in pink; hereafter referred to as CES triplets. As shown in the example, \textit{``the bombing created panic among villagers''}, illustrates that the event ``bombing'' caused the event ``panic among villagers'' termed as \textit{effect}. The linkage among the cause and effect, i.e., the word ``created'', is termed as \textit{signal} and can be expressed explicitly or implicitly. 
Automatically detecting and extracting causality relations plays a vital role in many natural language processing (NLP) works to tackle inference and understanding \cite{dunietz2020test,fajcik-etal-2020-fit,jo2021classifying,feder2021causal}. It has applications in various down-streaming NLP tasks, namely, causal question-answering generation, explaining social media behavior, political phenomena, effective education, and gender bias in the research community \cite{tan2014effect,wood2018challenges,sridhar2019estimating,veitch2020adapting,zhang2020quantifying,feder2021causalm}. 
%It plays a vital role as in many NLP task that aim to tackle inference and understanding 
%\begin{example}
%The bombing created panic among villagers 
%\end{example}
\def\ourunderline#1{\underline{\vphantom{[j}#1}} % so that the underlines are at the same depth
\begin{figure}[t]
\scriptsize
\begin{tabular}{|p{0.45\textwidth}|}
\hline
\vspace{-2px}
\textbf{(A) Casual segment:}\\
The treating doctor said \colorbox{green}{\dotuline{Sangram lost around 5kgs}} \colorbox{pink}{\dashuline{due to}} \colorbox{yellow}{\uwave{the hunger strike}}.\\
\colorbox{yellow}{\uwave{The bombing}} \colorbox{pink}{\dashuline{created}} \colorbox{green}{\dotuline{panic among villagers}}.\\
\colorbox{yellow}{\uwave{Dissatified with the package}}, \colorbox{green}{\dotuline{workers staged an all-night sit-in}}.\\
\hline
\textbf{(B) Non Casual Segment:}
Thus \colorbox{green}{\dotuline{we too joined the sloganerring}}.\\
The alliance claimed 4,000 took part last year.\\
\hline
% \textbf{(C) Our Model results :} \colorbox{yellow}{The end of the hunger strike}\colorbox{pink}{came as a} \colorbox{green}{relief to the jail administration}.\\
% \colorbox{yellow}{The worker had embarked on a wildcat strike} \colorbox{pink}{demanding}\colorbox{green}{better working condition}\\
% \colorbox{yellow}{Thirty-four people mostly striking mineworkers were shot} dead in \colorbox{green}{a clash with police}. \\
%\hline
\end{tabular}
\caption{Examples from the Causal News Corpus, causes are in \uwave{yellow}, effects in \dotuline{green}, and signals in \dashuline{pink}. If a sentence has both --- cause and effect --- it is referred to as casual (A), otherwise, as non-casual (B).}
\label{Figure:example}

\end{figure}

%Author citation \cite{tan-EtAl:2022:LREC}. \todo{You can also promise that our code will be released online.}

In this paper, we describe our methodology for CASE-2022 cause-effect-signal span detection shared task (Subtask 2). Overall, our main contributions are listed below:

\begin{enumerate}
    \item We show that cause-effect-signal spans can be extracted by a simple pre-trained generative seq2seq model trained on just 160 instances.
    \item We develop a method for extracting all causal triplets from the sentence in an iterative manner.
    \item We investigate how language models deal with the causal order of the cause and effect spans to answer the research question \emph{``should cause be identified first, and only then effect, or vice-versa?''}.
    \item We show that an efficient F1 best-substring matching algorithm, known for question answering, can be applied to deal with rare cases when a language model (LM) does not generate part of the input sequence.
\end{enumerate}

\section{Related Work}
The problem of causality extraction from text is a challenging task as it requires semantic understanding and contextual knowledge. There were many attempts in the domain of linguistics for corpora creation for event extraction but with limited size such as CausalTimeBank (CTB) \cite{mirza2014annotating} from news with 318 pairs, CaTeRS \cite{mostafazadeh2016caters} from short stories with 488 casual links, EventStoryLine \cite{caselli2017event} from online news articles with 1,770 casual event pairs, semantic relation corpora PDTB-3 \cite{webber2019penn} with over 7, 000 causal relations and CNC corpus \cite{tan-EtAl:2022:LREC, Fiona2022} with 1,957 casual events with multiple event pairs. %Compared to the previous dataset, it is annotated manually by experts by incorporating casual linguistic constructions.
Compared to previous datasets, CNC differs by focusing on event sentences, accepting arguments which does not need to form a clause, and not limiting itself to pre-defined list of connectives, but instead including causal examples in more varied linguistic constructions.
The previous work in this domain can be broadly classified into knowledge-based approaches, statistical ML, and deep-learning-based approaches. The knowledge-based approach uses linguistic patterns by predefining hand-crafted or keywords \cite{garcia1997coatis,khoo2000extracting,radinsky2012learning,beamer2008automatic,girju2009classification,ittoo2013minimally,kang2014knowledge,khoo1998automatic,bui2010extracting}. 

%Statistical techniques \cite{girju2003automatic} use third-party NLP tools such Wordnet \cite{miller-1994-wordnet} to generate a set of features for a given collection of labeled data, and ML algorithms (support vector machines, maximum entropy, naive bayes, and logistic regression) to perform the relevance classification.
Statistical techniques \cite{girju2003automatic,do2011minimally} rely on building probabilistic models over features extracted via third-party NLP tools such as Wordnet \cite{miller-1994-wordnet}. Deep-learning techniques map words and features into low-dimensional dense vectors, which may alleviate the feature sparsity problem. The most frequent used sequence to sequence models are feed-forward network \cite{ponti2017event}, long short-term memory networks \cite{kruengkrai2017improving,dasgupta2018automatic,martinez2017neural} convolutional neural networks \cite{jin2020inter,kruengkrai2017improving,wang2016relation}, recurrent neural networks \cite{yao2019graph}, gated recurrent units \cite{chen2016implicit} which embed semantic and syntactic information in local consecutive word sequences \cite{yao2019graph}. Later unsupervised training model such as BERT~\cite{devlin2018bert,sun2019deep}, RoBERTa~\cite{becquin2020gbe}, graph convolution network~\cite{zhang2018graph}, graph attention networks and joint model for entity relation extraction~\cite{li2017neural,wang2020two,zhao2021modeling,bekoulis2018adversarial}.

In this work, we base our model on T5 \cite{raffel2020exploring}, a sequence-to-sequence transformer model, pre-trained on a mixture of denoising objective and ~25 supervised tasks such as machine translation, linguistic acceptability, abstractive summarization or question answering. The unsupervised denoising objective randomly replaces spans of the input with different mask tokens, and generates contents of these masked spans prefixed with these special mask tokens. Furthermore, our work shares similarities with pointer-network \cite{vinyals2015pointer} based generative framework for various NER subtasks introduced by \citet{yan2021unified}. Contrastively, our work is more adapted to low-resource scenarios, as no extra parameters were added to our system, at the cost of errors, which can happen in the postprocessing matching step.
%There has been limited work for span-based casualty extraction \cite{becquin2020gbe,mariko2020financial}, and to the best of our knowledge, no work for the extraction of cause-effect-span detection in the news domain.

\section{Problem Description}
CASE-2022 shared task challenge \cite{tan-etal-2022-event} aimed for event causality identification, and extraction in casual news corpus \cite{tan-EtAl:2022:LREC}. It comprised of two subtasks, namely casual event classification (Subtask 1) and cause-effect-signal span detection (Subtask 2)\footnote{We participated in both subtasks, but report on Subtask 2 in this paper. For Subtask 1, we refer reader to our standalone publication \cite{idiap_subtaskA}.}. 
Subtask 2 aims on extracting the spans corresponding to cause-effect-signal (CES) triplets, as shown in Figure \ref{Figure:example}. We trained a generative seq2seq model to address this challenge and extracted the CES triplets using an iterative procedure (see Section \ref{sec:vanilla_model}).

% \begin{table*}
% \scriptsize
% \centering
% \begin{tabular}{lllllll}
% \hline
% \multicolumn{1}{l}{}      & \multicolumn{3}{c}{\textbf{Train phase}}                                                          & \multicolumn{3}{c}{\textbf{Test phase}}                                                           \\ \hline
% \multicolumn{1}{l}{}       & \multicolumn{1}{l}{\#Sentences} & \multicolumn{1}{l}{\#Relations} & \multicolumn{1}{l}{Gold labels} & \multicolumn{1}{l}{\#Sentences} & \multicolumn{1}{l}{\#Relations} & \multicolumn{1}{l}{Gold labels} \\ \cline{2-7} 
% train/dev                   & 160                            & 183                            & yes                              & 15                             & 18                             & yes                              \\
% grouped & 2925                           &                                & partial                          & 323                            &                                & partial                          \\
% %dev\_text/test\_text        & 323                            &                                & no                               & 311                            &                                & no                              \\\hline
% \end{tabular}

% \end{table*}
\begin{table}[t]
    \centering
    \resizebox{\linewidth}{!}{\begin{tabular}{llll}
    \toprule
    Split & \#Sentences & \#Relations & \#Signals  \\
    \midrule
    Train & 160         & 183         & 118 (64\%) \\
    Dev   & 15          & 18          & 10 (56\%)  \\
    Test  & 89           & 119         & 98 (82\%) \\
    \bottomrule
\end{tabular}
}%
    \caption{Dataset statistics. See text for details.}
    \label{tab:dataset_stats}
\end{table}
% \todo{Replace this paragraph with Dataset statistics Table and its description}
% The dataset is divided into : data for training and testing phase. The training phase data contains train, train-grouped and dev\_text set. The train set contains (160 sentences and 183 relations) and train\_grouped has 2925 sentences with gold labels. The dev text set contains 323 sentences without the gold labels. While the dataset for test phase, contains dev, dev\_grouped, test\_text. The dev contains 15 sentences, 18 relations with gold labels and dev\_grouped contains 323 sentences with partial gold labels while the test\_text contains 311 sentences without the gold labels.
The dataset statistics are presented in Table \ref{tab:dataset_stats}. The number of total sentences is given by the column \emph{\#Sentences}, whereas a total number of CES triplets is in column \emph{\#Relations}. Column \emph{\#Signals} shows how many signal annotations were present in the total number of CES triplets.

%... Each example contained up to 4 CES triplets in training data....
%... \todo{Define CES triplet as Cause-Effect-Signal triplet}...
\section{Methodology}
\subsection{Language Model Training}
\label{sec:vanilla_model}
We utilize T5 \cite{raffel2020exploring}, a pre-trained autoregressive transformer-based language model trained on a mixture of unsupervised and supervised tasks that require language understanding. The model is conditioned $n\times3$ times for each example, as there can be $n$ CES triplets in one sentence (up to $n=4$ triplets in training data). Each time, we condition the language model 3 times for every example and its corresponding CES triplet, generating a different triplet component (cause, effect, and signal) to learn to generate the entire CES triplet. As these triplets are unordered, we uniformly sample a random path among them (e.g., 2-3-1-4, for sample with four triplets) during training. We only train with as many triplets, as available in the training data. We now describe the input format, further illustrated in Appendix \ref{sec:example_of_inputs}.

Firstly, the model's encoder is conditioned with sentence tokens {\small\texttt{<sentence>}} followed by the history of already generated CES triplets for this example (empty if there was none) as
\begin{myquote} 
{\small\texttt{<sentence> \_history : <history>}}.
\end{myquote}
The history is always prepended with {\small\texttt{\_history:}} tokens. The content of the history are the already generated triplets. Each part of the triplet is prepended with its corresponding {\small\texttt{\_cause:}}, or {\small\texttt{\_effect:}}, or {\small\texttt{\_signal:}} sequence.
Concurrently, model's decoder is prefixed with {\small\texttt{\_cause:}} sequence. In this case, the probability of cause sequence is maximized.

Secondly, the model is conditioned with sentence tokens {\small\texttt{<sentence>}} and cause tokens {\small\texttt{<cause>}}, prepended with {\small\texttt{\_cause:}} token as 
\begin{myquote} 
{\small\texttt{<sentence> \_cause : <cause> \_history : <history>}}.
\end{myquote}
This time, the decoder is prompted with {\small\texttt{\_effect:}} prefix, and the probability of effect sequence is maximized.

Thirdly, the model is conditioned with sentence tokens {\small\texttt{<sentence>}}, cause tokens {\small\texttt{<cause>}}, and effect tokens {\small\texttt{<effect>}} with {\small\texttt{\_effect:}} token prepended as  
\begin{myquote} 
{\small\texttt{<sentence> \_cause : <cause> \_effect : <effect> \_history : <history>}}.
\end{myquote}
Analogically, decoder is prompted with {\small\texttt{\_signal:}} prefix and probability of signal sequence is maximized. As the signal might not always be part of the CES triplet, we let the model generate {\small\texttt{\_empty}} token in these cases.

\subsection{Experimental Details}
\label{sec:expdetails}
We use cross-entropy (CE) loss to train the T5. We firstly average CE loss over tokens, then over inputs per example (for all CES triplets), and then across mini-batch. We use greedy search to generate the sequences. In inference time, we always generate 4 CES triplets for each sentence, as that is the maximum we observed in the training data. 

As we don't constrain the decoding, the generated sequence does not have to match certain sub-string in the input. However, the extractive task requires inserting tags around a cause, effect, or signal span inside the input sentence. Therefore we map the generated sequences back to the input sentence via F1 matching. In particular, for each generated sequence, we find the most similar substring in the input, where the similarity is measured via token-level F1 score. We utilize an efficient F1 matching technique, which prunes out a significant part of the search space, presented in the Appendix C.1 of \citet{fajcik-etal-2021-r2-d2}\footnote{Implemented at \url{https://shorturl.at/kxEVW}.}. We base our implementation on PyTorch \cite{NEURIPS2019_9015}, Transformers \cite{wolf-etal-2020-transformers} libraries and use AdamW \cite{loshchilov2017decoupled} for optimization. We tune hyperparameters via HyperOpt \cite{bergstra2015hyperopt} and report the exact hyperparameters in Appendix \ref{sec:hyperparameters}.

\subsection{Evaluation Metrics}
In this section, we describe the metrics we used to evaluate the system. %% Commented out, as we dont have space%%
% Given predicted sentences,\textit{``Violence broke out when <ARG1> DYFI members protested </ARG1> <ARG0> <SIG0>against</SIG0> a Youth Congress meeting being held at the junction </ARG0>''}
% for a reference annotated sentence, 
% \textit{``<ARG1> Violence broke out </ARG1> when <ARG0> DYFI members protested </ARG0> against a Youth Congress meeting being held at the junction''} we  calculate the F1 score for cause, effect and signal. The pairing of prediction and reference that returns the highest overall F1 score is kept as final score. 
\vspace{-1ex}
\begin{description}[style=unboxed,leftmargin=0em,listparindent=\parindent]
    \setlength\parskip{0em}
\item \textbf{F1:} F1 score was the official main evaluation metric in the challenge. It is computed over B, and I tags in sequence following the BIO tagging scheme for every example and every CES triplet component separately, using \texttt{seqeval}\footnote{\url{https://github.com/chakki-works/seqeval}.}. The F1 is then averaged firstly across dataset examples, obtaining F1 for each component (\textit{{Cause F1}}, \textit{{Effect F1}}, \textit{{Signal F1}}). \textit{{Overall F1}} is computed as a weighted average of component examples by their frequency. 
\item \textbf{CE:} is an average token cross-entropy, computed as described in Section \ref{sec:expdetails}. 
\item \textbf{ES Acc:} is an empty-signal accuracy, i.e., an accuracy of the model predicting no signal span in the CES triplet when given golden cause and effect.
\end{description}

\subsection{Baseline Model}
As a baseline model, we used the CASE-2022 organizers' provided model for Subtask 2: a random generator that uniformly samples a cause, effect, and signal spans\footnote{Available at \url{https://shorturl.at/msY04}.} from the sentence. This baseline guarantees the cause and the effect do not overlap.

\section{Results \& Discussion}
\begin{table}
    \centering
    \small
    \scalebox{1.0}{\begin{tabular}{l@{~~}c@{~~}c@{~~}c@{~~}c@{~~}c@{~~}c}
\toprule
\textbf{System}                        & \textbf{CE} & \textbf{Cause} & \textbf{Effect} & \textbf{Signal}    & \textbf{Overall} \\
\midrule
\multicolumn{1}{l|@{~~}}{Baseline}          & -           & -                 & -                  & -                     &  2.2                \\
\multicolumn{1}{l|@{~~}}{T5-NoHistory}      & .181        & -                 & -                  & -                     & 67.7$\pm$2          \\
\multicolumn{1}{l|@{~~}}{T5-ECS}            & .168        & 75.9$\pm$5        & 71.3$\pm$4         & 76.1$\pm$5            & 73,5$\pm$2          \\
\multicolumn{1}{l|@{~~}}{T5-CES}            & .183        & 81.0$\pm$4        & 67.8$\pm$2         & 66.7$\pm$5            & 73.0$\pm$2          \\
\multicolumn{1}{l|@{~~}}{T5-CES$_{LARGE}$}  & .159        & 73.5$\pm$8        & 74.1$\pm$4         & 77.2$\pm$7            & 74.8$\pm$2          \\
\bottomrule
\end{tabular}}%
    \caption{Main results, in terms of Cross-Entropy (CE) and F1, with $\pm$ standard deviations on dev data.}
    \label{tab:main_results}
\end{table}

We now report the results obtained from averaging at least ten measured performances from 10 checkpoints trained with different seeds\footnote{Dev set predictions from our best t5-base model are available at \url{https://shorturl.at/bjVZ9}.}. We studied 4 different variants of our system. System T5-CES is our vanilla model described in \ref{sec:vanilla_model}, based on T5-base. System T5-CES$_{LARGE}$ is the same model based on T5-large. Unlike T5-CES, system \mbox{T5-ECS} reverses the generation order by generating the first effect and cause, followed by the signal (assuming causal order \emph{effect$\rightarrow$cause$\rightarrow$signal}, hence the suffix ECS). Lastly, we studied the effect of conditioning the model on the history of already generated triplets. We remove the history from the input at all times in training and predict the four identical CES triplets for each example in test time.
Our ablated results are available in Table \ref{tab:main_results}.

Firstly, the model with no history at input performs significantly worse, validating our hypothesis that the model can learn to decrease the probability of the triplets already contained within the input, even from just 160 samples. 
Secondly, we observed a general trend that \textit{in the Cause F1 T5-CES outperforms T5-ECS} and \textit{in Effect F1, T5-ECS outperforms T5-CES}. This leads to the hypothesis that whichever part of the triplet, cause or effect, is generated first, the language model performs better in its case.
Thirdly, we observed that the large model achieved the best results on average. It also achieved our best single-checkpoint performance on the dev set (78.3 Overall F1).
However, given the sample size of the dev set, the differences between T5-CES, T5-ECS, and T5-CES$_{LARGE}$ can hardly be deemed significant.

Next we present our results on the test set in Table \ref{tab:results_test}. We submitted checkpoints with the best overall F1 score on the dev set (Dev F1) to the leaderboard while varying the model types. We observed a significant drop in performance on the test data. As the annotation on the test data is not released at the time of writing, the causes of this performance drop remain unknown. We hypothesize it could have been caused by a covariate shift in the test data, as supported by \#Signals statistics in Table \ref{tab:dataset_stats}.

Additionally, we include extra statistics (Dev$_0$~F1, Dev$_1$~F1, Dev~ES~Acc) for our best checkpoints. We expected the performance on the dev subset with two triplets (Dev$_2$ F1) per example to be worse than on the dev subset with one triplet per sentence (Dev$_1$ F1). Performance-wise this does not always seem to be the case. Upon manual analysis, we found that the model often failed in the second round of triplet extraction. We found 2 LM hallucinations out of 18 dev samples in the second generation round.

\begin{table}[t]
    \centering
    %\resizebox{\linewidth}{!}{\input{tables/results_test}}%
    \small
    \scalebox{1.0}{\begin{tabular}{l@{~~~~}c@{~~~~}c@{~~~~}c@{~~~~}c@{~~~~}c@{~~~~}c}
\toprule
\multirow{2}{*}{\textbf{System}}  & \textbf{Dev} & \textbf{Dev$_{1}$} & \textbf{Dev$_{2}$} & \textbf{Dev}   & \textbf{Test}\\ 
  & \textbf{F1} & \textbf{F1} & \textbf{F1} & \textbf{ES Acc}   & \textbf{F1}    \\ 
\midrule
\multicolumn{1}{l|@{~~}}{T5-ECS}            & 77.7            &   80.9              &  71.1               &  82                   & 43.4                \\
\multicolumn{1}{l|@{~~}}{T5-CES$_{LARGE}$}  & 78.3            &   77.4              &  80.0               &  70                   & 43.7                \\
\multicolumn{1}{l|@{~~}}{T5-CES}            & 77.5            &   79.6              &  73.3               &  70                   & \textbf{48.8}  \\
\bottomrule
\end{tabular}}%
    \caption{Top checkpoints submitted to the leaderboard.}
    \label{tab:results_test}
\end{table}
\section{Inference Speed}
Measuring the inference speed on test set, we used Intel i5-based 2080Ti GPU workstation. The inference of 4 CES triplets without postprocessing per 1 sentence example took $1.46$ seconds on average. The postprocessing runtime was negligible, taking $0.025$ seconds per sentence example on average.

\section{Conclusion}
In this work, we have analyzed our CASE-2022 2nd place submissions on Subtask 2. We showed that a generative model could extract cause-effect-signal triplets at the competitive level using just 160 annotated samples. We investigated causal assumptions about the generation order of cause and effect to answer the research question \emph{``should cause be identified first, and only then effect, or vice-versa?''} and found that while the Overall F1 won't change significantly, whichever component was generated first achieved better performance on average (Cause first achieved better Cause-F1, and Effect first Effect-F1 respectively). Finally, we showed the F1 difference between the dev subset with 1 or 2 causal triplets per sentence is negligible.

\section*{Acknowledgements}
This work was supported by CRiTERIA, an EU project, funded under the Horizon 2020 programme, grant agreement no. 101021866 and the Ministry of Education, Youth and Sports of the Czech Republic through the e-INFRA CZ (ID:90140). Esaú Villatoro-Tello, was supported partially by Idiap, SNI CONACyT, and UAM-Cuajimalpa Mexico.

% \vfill\eject %TODO: remove after review!

\bibliography{custom}
\bibliographystyle{acl_natbib}

\appendix

\newpage
\section{Hyperparameters}
\label{sec:hyperparameters}
In Table \ref{tab:hyperparameters}, we report the exact hyperparameters used when fine-tuning T5. Warmup proportion, weight decay, and dropouts are in the (0,1) range (for instance, .4719 means 47.19\%).

\begin{table}
    \centering
    \scalebox{1.0}{\begin{tabular}{l|l}
\hline
\textbf{Hyperparameter} & \textbf{Value}             \\ \hline
learning rate           & .0002                      \\
hidden dropout          & .1436                      \\
attention dropout       & .4719                      \\
weight decay            & .0214                      \\
minibatch size          & 8                          \\
warmup proportion       & .1570                      \\
scheduler               & constant (no lr decrease)  \\
max steps               & 10,000                     \\
max gradient norm       & 1                      \\\hline   
\end{tabular}}%
    \caption{Hyperparameter setting used in this work.}
    \label{tab:hyperparameters}
\end{table}

\section{Example of Inputs}
\label{sec:example_of_inputs}
\begin{figure*}
    \centering
    \scalebox{1.1}{\includegraphics{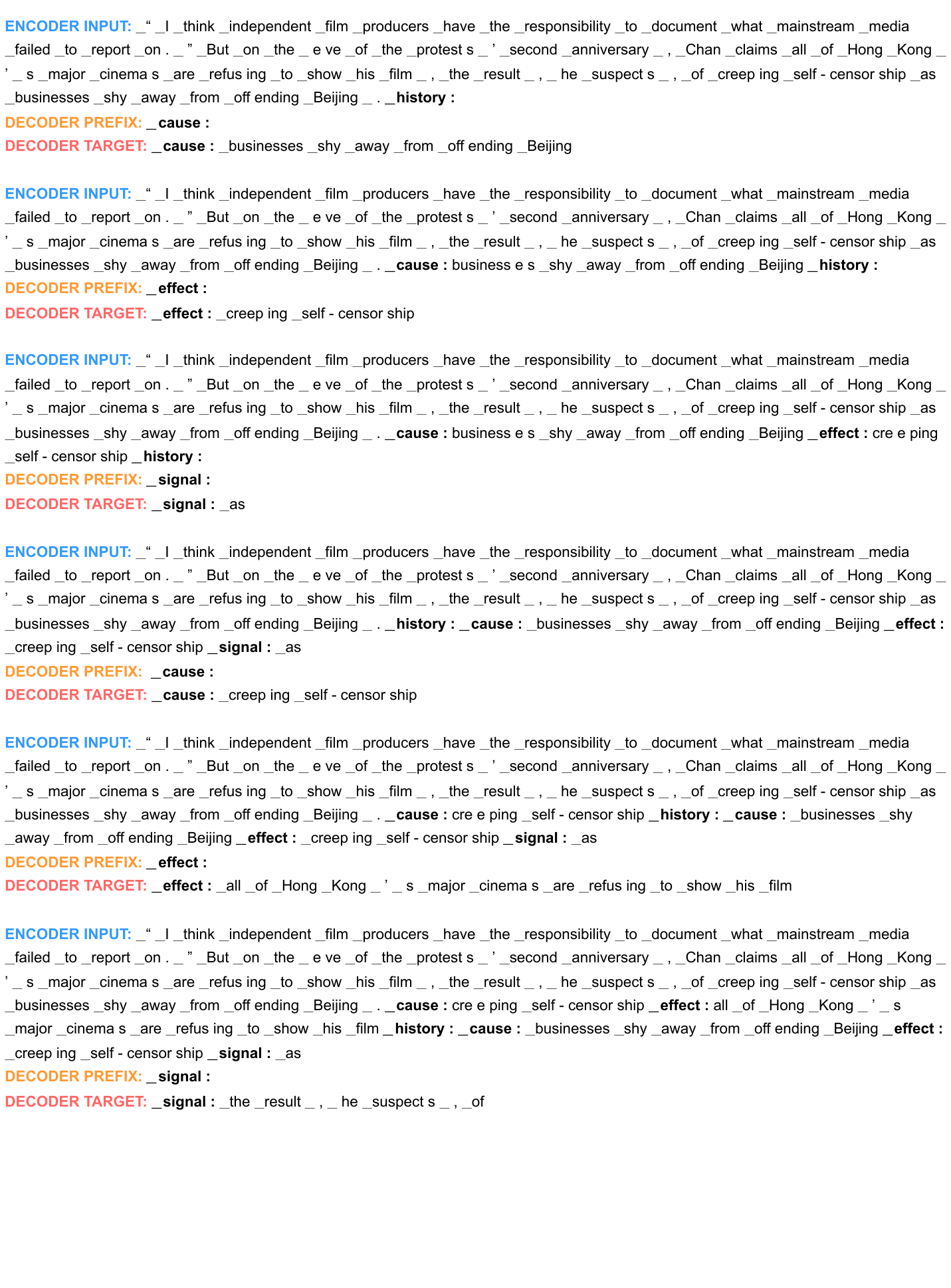}}
    \caption{Example of tokenized inputs for a sentence with two annotated CES triplets. Phrases "ENCODER INPUT". "DECODER PREFIX" and "DECODER TARGET" are not parts of the input, and are included for illustrative purposes only. Special sequences (\texttt{\_cause:, \_effect:, \_signal:, \_history:})  used between concatenated parts of the input are in bold.}
    \label{fig:example_of_inputs}
\end{figure*}

The input format and label format for a single training example, a sentence with 2 CES triplets, are illustrated in Figure. \ref{fig:example_of_inputs}.
\label{sec:appendix}

\end{document}